\documentclass[aip, floatfix, jcp, reprint, longbibliography]{revtex4-1}

\usepackage{enumitem}
\usepackage{graphicx}          
\usepackage{amsmath}    
\usepackage{amssymb}    
\usepackage{amsthm}
\usepackage[]{algorithm}
\usepackage{algpseudocode}
\usepackage{enumitem}
\usepackage{graphicx}   
\usepackage{subfigure}  
\usepackage{hyperref}   
\usepackage[capitalise]{cleveref}   
\usepackage{appendix}

\begin{document}

\title{Simultaneous coherent structure coloring facilitates interpretable clustering of scientific data by amplifying dissimilarity}

\author{Brooke E. Husic}
\email{bhusic@stanford.edu}
\affiliation{Department of Chemistry, Stanford University, Stanford, California, United States of America}
\author{Kristy L. Schlueter-Kuck}
\affiliation{Department of Mechanical Engineering, Stanford University, Stanford, California, United States of America}
\author{John O. Dabiri}
\email{jodabiri@stanford.edu}
\affiliation{Department of Mechanical Engineering, Stanford University, Stanford, California, United States of America}
\affiliation{Department of Civil and Environmental Engineering, Stanford University, Stanford, California, United States of America}

\begin{abstract}
The clustering of data into physically meaningful subsets often requires assumptions regarding the number, size, or shape of the subgroups. Here, we present a new method, simultaneous coherent structure coloring (sCSC), which accomplishes the task of unsupervised clustering without \textit{a priori} guidance regarding the underlying structure of the data.
sCSC performs a sequence of binary splittings on the dataset such that the most dissimilar data points are required to be in separate clusters. To achieve this, we obtain a set of orthogonal coordinates along which dissimilarity in the dataset is maximized from a generalized eigenvalue problem based on the pairwise dissimilarity between the data points to be clustered.
This sequence of bifurcations produces a binary tree representation of the system, from which the number of clusters in the data and their interrelationships naturally emerge.
To illustrate the effectiveness of the method in the absence of \textit{a priori} assumptions, we apply it to three exemplary problems in fluid dynamics.
Then, we illustrate its capacity for interpretability using a high-dimensional protein folding simulation dataset.
While we restrict our examples to dynamical physical systems in this work, we anticipate straightforward translation to other fields where existing analysis tools require \textit{ad hoc} assumptions on the data structure, lack the interpretability of the present method, or in which the underlying processes are less accessible, such as genomics and neuroscience.
\end{abstract}

\maketitle

\section*{Introduction}

Modern science increasingly leverages machine learning on large datasets in the sciences, from electronic structure~\cite{wu2018moleculenet} to whole genome sequences~\cite{kavvas2018machine} to distributed ocean sensor measurements~\cite{FroylandPadberg_Chaos15}.
Many of these datasets capture the dynamics of a system evolving in time, encoding trends with predictive power.
Analyzing these datasets using a statistically robust and interpretable framework is a longstanding challenge that often involves clustering, or the unsupervised learning of coherent groups within the dataset. 

Clustering is a notoriously challenging problem which, unlike supervised learning, features no direct measure of model success or validity and often requires heuristic assessments of effectiveness~\cite{friedman2001elements}.
Thus, many classes of clustering algorithms have been developed for different problems. Some commonly used techniques include partition-based methods such as $k$-means~\cite{Macqueen_MSP67}, or their fuzzy counterparts~\cite{Dunn_JCybern73}; density-based methods such as DBSCAN~\cite{Ester_KDD96}; and connectivity-based methods such as divisive and agglomerative hierarchical clustering~\cite{NewmanGirvan_PRE04,Kaufman_Book09}.

Each of the aforementioned methods exhibits drawbacks with respect to \textit{a priori} assumptions and algorithmic limitations.
For example, partition-based clustering such as $k$-means requires the modeler to prescribe the number of partitions in a dataset before constructing the model.
If multiple results are obtained from different values of $k$, these results are not interrelated; similarly, the model cannot be used to determine relationships between the $k$ clusters of a single model.
While connectivity-based methods feature interrelated clusters, these also require the determination of where to cut the corresponding dendrogram to obtain the clustering result.
Although density-based methods do not require \textit{a priori} or \textit{a posteriori} determination of the number of clusters to use, these methods are generally not robust to datasets containing a range of cluster densities~\cite{Ali_ICIET10}.

Here, we present a new method, simultaneous coherent structure coloring (sCSC), which minimizes the assumptions required in an unsupervised clustering task.
sCSC focuses solely on the efficient separation of the most dissimilar states in the system, resulting in a quantitative structure that automatically captures the clusters in the dataset and their interrelationships without \textit{a priori} knowledge of the system. The method is demonstrated for simulated and empirical systems of fluid and molecular dynamics, and its straightforward extension to other types of data is discussed.

\section*{Background}

The use of clustering for data analysis is ubiquitous. However, our motivation emerged from research on the identification of coherent structures from fluid dynamics. 
A variety of mathematical frameworks have been developed to identify coherent structures.
The broad class of Lagrangian methods has been developed to describe flows that are unsteady (i.e., not well-summarized by instantaneous snapshots) in a way that is not dependent on their frame of reference (i.e., may not contain velocity or acceleration data). Two recent reviews of Lagrangian methods for the detection of coherent structures in fluids can be found in Refs.~\onlinecite{AllshousePeacock_Chaos15} and~\onlinecite{Hadjighasemetal_Chaos17}.

Existing algorithms for coherent structure analysis that involve clustering exhibit various limitations. The fuzzy $c$-means approach presented in Ref.~\onlinecite{FroylandPadberg_Chaos15}, for example, introduces a dynamic distance between particle trajectories, but ultimately requires the choice of $c$, i.e.~how many clusters to use.
To avoid explicitly choosing the number of coherent structures, a spectral clustering method was introduced in Ref.~\onlinecite{Hadjighasemetal_PRE16}, which utilizes the spectral gap in the graph Laplacian to determine the number of coherent structures. However, it was subsequently shown in Ref.~\onlinecite{SchlueterKuck_JFM17} that such a gap is only robust when the number of trajectories used exceeds $10^3$.

The method of coherent structure coloring (CSC), introduced in Refs.~\onlinecite{SchlueterKuck_JFM17} and~\onlinecite{SchlueterKuck_Chaos17}, was designed to address these and other limitations of clustering algorithms for coherent structure determination based on trajectories of particle spatial coordinates. In this work, we extend CSC in the context of its own limitations, as described below.

While we restrict the focus of the rest of the paper to clustering, this is not the only way to identify coherent structures from frame-independent particle trajectories.
Over the past two decades, both the fluid and molecular dynamics communities have developed methods to identify ``almost-invariant'' sets through data-driven approximations to the Perron-Frobenius operator and its adjoint, the Koopman operator. The former, also referred to as the transfer or transition operator, propagates probability densities forward in time, whereas the latter propagates observables~\cite{klus2016numerical}.

For fluid systems, Dellnitz and Junge~\cite{dellnitz1999approximation}, Froyland and Dellnitz~\cite{froyland2003detecting}, and Mezi{\'c}~\cite{mezic2005spectral} used finite approximations to the Perron-Frobenius eigenfunctions to divide the space occupied by a dynamical system into almost-invariant sets and almost-cycles.
At the same time, Sch{\"u}tte \textit{et al}.~\cite{schutte1999direct} and Deuflhard \textit{et al}.~\cite{Deuflhard_LAA00} introduced the use of approximations to the same operator, under a reversibility constraint, to determine the ``metastable'' states of molecular systems simulated on the atomic level.

A decade later, researchers in their respective fields independently determined equivalent algorithms for optimizing the estimation of the approximated eigenfunctions of the Perron-Frobenius and Koopman operators~\cite{Noe_MMS13,williams2015data,wu2017variational}.
In both cases, linear models are generated using a data-driven, objective protocol to model highly nonlinear dynamics, where the eigenvectors and eigenvalues can be used to identify coherent sets.

In fact, in our final example in the current study, we utilize both types of methods on a molecular dynamics simulation dataset. We first create an optimized model that approximates the Perron-Frobenius operator~\cite{Noe_MMS13}, which is difficult to visualize due to its high dimensionality. Then, to reduce our model to a visualizable and interpretable coarse-grained model, we use the clustering method presented in this work to identify the major coherent structures.

\section*{Simultaneous coherent structure coloring (\lowercase{s}CSC)}

\subsection*{Coherent structure coloring theory}

Many datasets we wish to explore in the physical sciences are generated by complex dynamical systems that exhibit instabilities and chaos.
A key consequence of these processes is that states of the system (e.g.~fluid particle trajectories or protein conformations) that are proximal but belonging to different coherent sets will separate exponentially faster as the system evolves than states belonging to the same cluster~\cite{Guckenheimer_Book83,HallerYuan_PhysicaD00}.

On this basis, we previously hypothesized that these complex datasets can be clustered more robustly and effectively by amplifying state differences rather than state similarity~\cite{SchlueterKuck_JFM17,SchlueterKuck_Chaos17}.
The rationale for this approach is that the exponential separation of dissimilar states can provide more sensitive detection of clusters than a focus on state similarity, the latter requiring longer observation to become apparent~\cite{Guckenheimer_Book83,HallerYuan_PhysicaD00}.
In other words, we aim to identify coherent clusters indirectly, by prioritizing the separation of states with greatest dissimilarity and confidently ruling out the possibility of their membership in the same cluster.
Those states that remain together after the separation process will subsequently emerge as belonging to the same cluster. 

To amplify the dissimilarity between states, we solve an optimization problem to maximize a figure of merit $z$ that quantifies total state dissimilarity in the dataset. 
Specifically, this figure of merit depends on a scalar value $x_i$ assigned to each state $i$ in the system, where the squared difference in the scalar value assigned to each of pair states (e.g.~$(x_1 - x_2)^2$ for states 1 and 2) is weighted by a measure of their dissimilarity. Formally, the clustering parameter $z$ is given by

\begin{equation} \label{eq:z}
z \equiv \frac{1}{2} \sum_{i}^{n}\sum_{j}^{n} (x_i - x_j)^2 a_{ij},
\end{equation}

\noindent{}where the summations of $i$ and $j$ are each taken over the full set of $n$ states to be clustered, and $a_{ij}$ is an element of the adjacency matrix $A$ containing the pairwise dissimilarity between states $i$ and $j$.
The construction of this matrix requires the calculation of ${n \choose 2} = (n-1)n/2$ adjacency values.
Example definitions of the (symmetric) pairwise dissimilarity can include the standard deviation for comparison of time-dependent signals, or the Jensen-Shannon divergence for comparison of probability distributions~\cite{Lin_IEEE91,Husic_Draft17}.
Both definitions represent measures of dissimilarity, where identical data points receive $a_{ij}=0$. Thus, assuming all data points are unique, the matrix $A$ will be dense.

Given the adjacency matrix $A$, we seek to find state assignments $x_i$ that will maximize $z$, subject to the constraint that the magnitude of the $n \times 1$ vector $X$ containing the $n$ scalar values $x_i$ must remain finite (e.g.~to avoid the trivial case that maximizes $z$ for $x_1 = \infty$ and $x_2 = -\infty$). It is straightforward to show that the constrained optimization of equation \ref{eq:z} with finite $X$ can be written as the generalized eigenvalue problem~\cite{Hall_MS70}:

\begin{equation} \label{eq:geneig}
L X = \lambda D X,
\end{equation}

\noindent{}where $D$ is a diagonal matrix with entries equal to the row-sums of the adjacency matrix, i.e.~$\sum_{j} a_{ij}$ for each row $i$, and $L = D - A$ is the graph Laplacian. This maximization is expressed using the Lagrangian form; see~\cite{SchlueterKuck_JFM17} for more details.

Each of the $n$ eigenvectors $X_n$ of equation \ref{eq:geneig} represents a solution that assigns to each state a scalar value $x_i$ based on its dissimilarity to the other states in the system.
Those states with scalar assignments in each $X$ that are most dissimilar can be presumed to belong to different clusters of the data when the data is partitioned according to that particular solution of equation \ref{eq:z}.
The eigenvector $X_1$ associated with the maximum eigenvalue $\lambda_1$ contains the scalar assignments $x_i$ that maximize the figure of merit $z$. This can be considered the single most effective partitioning of the dataset.

Given the analogy between this approach and the problem of fuzzy graph coloring~\cite{Munoz_Omega05}, wherein the connected nodes of a graph with large weights are assigned the most disparate values, we call this method coherent structure coloring (CSC)~\cite{SchlueterKuck_JFM17}.
The technique has recently been demonstrated to successfully identify coherent eddies and jets associated with individual fluid particle trajectories in model geophysical flows~\cite{SchlueterKuck_Chaos17}.

\begin{figure*}[thb!]
\centering
\includegraphics[width=\textwidth]{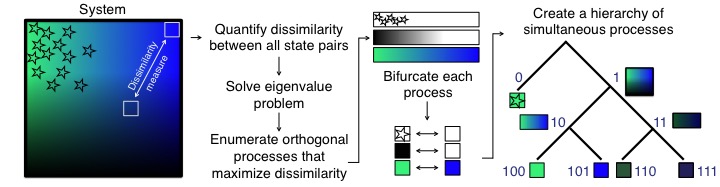}
\caption{Conceptual scheme illustrating the sCSC algorithm.
First, the dissimilarity between all pairs of states are tabulated in an adjacency matrix. For this example system, states are represented by a uniform grid of squares, two of which are illustrated in the left panel. The adjacency matrix is then used to solve an eigenvalue problem (equation~\eqref{eq:geneig}) that maximizes the dissimilarity measure. The solutions to the eigenvalue problem identify orthogonal processes in the system in order of their ability to separate the system; in this case, we have stars $\leftrightarrow$ no stars, bright $\leftrightarrow$ dark, and green $\leftrightarrow$ blue, which we have asserted are decreasingly effective in explaining dissimilarity in this notional system.
These three processes are bifurcated into two extremes (middle panel). Then, each state is encoded according to each bifurcation.
For the first orthogonal process (stars $\leftrightarrow$ no stars), we bifurcate the entire system.
For the next orthogonal process (bright $\leftrightarrow$ dark), we bifurcate the system separately and illustrate only states which become bifurcated along this division.
For example, there is no state that contains stars and is dark, so branch 0 of the corresponding dendrogram is not further bifurcated.
Finally, we bifurcate both branches 10 and 11 according to green or blue.}
\label{fig:scheme}
\end{figure*}

\subsection*{Simultaneous inclusion of multiple CSC solutions}

A key limitation of the original CSC method~\cite{SchlueterKuck_JFM17} is that it relies on only a single eigenvector associated with the largest eigenvalue of equation~\ref{eq:geneig}.
Hence, although multiple dimensions of dissimilarity are almost always present in real data, the method cannot simultaneously distinguish between multiple types of dissimilarity in a dataset.
Moreover, the method applied to individual fluid particle trajectories in a subsequent study required a subjectively defined threshold to calculate eigenspace distances~\cite{SchlueterKuck_Chaos17}, and it was shown to produce degenerate results for fluid particles in chaotic regions of the flow (cf.~Fig~7 in Ref.~\onlinecite{SchlueterKuck_Chaos17}).   

Importantly, because the adjacency matrix $A$ introduced in the previous section is real and symmetric, the remaining eigenvectors associated with lesser eigenvalues provide additional, linearly independent (i.e.~orthogonal) solutions for partitioning the data, albeit less effectively~\cite{Press_Book83}.
The key innovation of the present work is to use \textit{all} of the eigenvectors in a top-down fashion to simultaneously cluster the system states.

To perform sCSC, we begin with the most effective partition given by the eigenvector associated with the maximum eigenvalue, and proceed through the set of orthogonal eigenvectors in order of decreasing eigenvalue.
This approach simultaneously reveals the coherent sets of the system, and eliminates the subjective user intervention required in the previous method~\cite{SchlueterKuck_Chaos17}.

Given a dissimilarity measure and resulting eigenvector solutions, the simultaneous coherent structure coloring (sCSC) algorithm begins by assigning to each state in the system a binary membership based on its corresponding scalar value along each orthogonal coordinate direction.
A bifurcation is appropriate given that each one-dimensional coordinate has two extreme ends toward which the optimization of equation \ref{eq:z} pulls dissimilar states.

The states are bifurcated along each coordinate dimension by using agglomerative clustering with average linkage (although other linkages or splitting methods could be used for this step; see e.g. Ref.~\onlinecite{mullner2013fastcluster}, Table 1.) and assigning to each state a value of 0 or 1 based on its membership within either of the two largest clusters of the resulting dendrogram.
Each eigenvector contributes a separate bit to the binary code associated with each of the states in the system, with the leading bit corresponding to the largest eigenvalue and the remaining bits concatenated in decreasing order of their corresponding eigenvalues.
Though we suggest using a bifurcation in general, the method does not prohibit the division of each eigenvector coordinate into three or more discrete bins, thus creating a $k$-way splitting and associated base-$k$ codes.

For each subsequent eigenvector, the bifurcation is performed for all data points (i.e.~states), and each is assigned a 0 or a 1.
For the $k$th eigenvector bifurcation, this enables $2^k$ numerically possible clusters (Fig~\ref{fig:scheme}).
For example, the first splitting produces branches $0$ and $1$, and the second splitting enables the population of $2^2$ unique clusters by appending $0$ or $1$ to each branch of the existing binary code ($ \{ 00, 01, 10, 11 \} $).
However, it may be the case that the numerically possible branch $01$ is not occupied because there is no data point that receives both a label of $0$ in the first bifurcation and a label of $1$ in the second bifurcation.
Thus, we hypothesize that branch $0$ (and its only occupied split, branch $00$) evidences a coherent region of the data.
In this way, a natural stopping criterion emerges from unoccupied bit codes during the binary splitting.

The binary codes generated by the aforementioned process can be visualized in a dendrogram, with each branch pair connecting those states that differ only at the least significant bit of their binary code.
The length of each branch pair is a measure of the dissimilarity between the groups connected by the branches, and it corresponds to the value of the summation in equation~\eqref{eq:z} computed only over those states connected by the branches.
Bits for progressively smaller eigenvalues are included at progressively lower levels of the dendrogram.
The dissimilarity between the groups connected at lower levels therefore generally becomes smaller as well.
While in principle the sCSC dendrogram should naturally truncate when no further splits occur, large amounts of data points or statistical noise may lead to insignificant (i.e., low-$z$) clusters or explore a combinatorially unfavorable number of splits. In that case, one may choose to truncate the dendrogram after a certain number of eigenvectors according to visual inspection, or determine a cutoff based on the magnitude of $z$ or the eigenvalue.

As in standard divisive and hierarchical clustering methods, the clustering models produced with sCSC are dependent on the adjacency definition supplied by the user.
Because the adjacency matrix summarizes pairwise dissimilarities only, this has the benefit of not requiring adherence to the triangle inequality~\cite{Hansen1997}---in fact, the data points need not exist in a well-defined space at all.
However, with this flexibility comes the drawback that a poor dissimilarity metric may obscure patterns in the data.
The dissimilarity measures used in this study have been shown to be effective in previous studies~\cite{SchlueterKuck_JFM17,Husic_Draft17}, and in general may require domain-specific knowledge to determine for a given dataset.

In the next two sections, we apply sCSC to benchmark problems in fluid dynamics in order to demonstrate its effectiveness in identifying coherent structures in the absence of \textit{a priori} assumptions. 
Then, we demonstrate the use of sCSC to determine the number and shape of flow structures involved in vortex ring entrainment using data obtained from empirical measurements the laboratory.
Finally, to highlight the interpretability of the sCSC dendrogram for high-dimensional datasets, we use sCSC to visualize an interpretable representation of an atomistic protein folding simulation.
Finally, we discuss the relationship of this method to other unsupervised clustering methods, and the possibility of extending sCSC beyond physical dynamical systems.

\begin{figure*}[thb!]
\centering
\includegraphics[width=\textwidth]{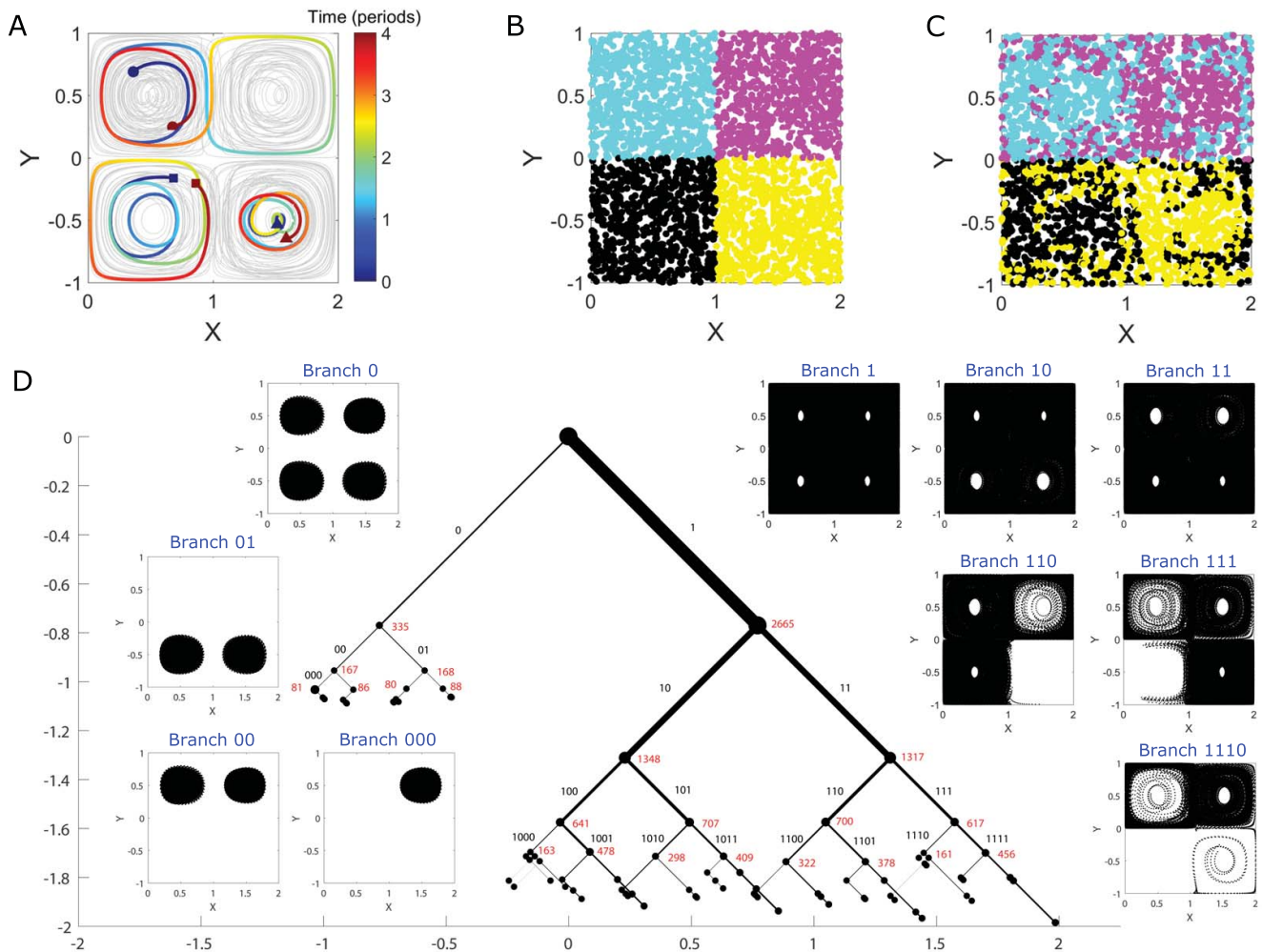}
\caption{Quadruple-eddy ocean flow model.
(A) Trajectories of 50 selected drifters randomly initialized in the flow (gray). The trajectories of 3 drifters are highlighted for 4 periods of horizontal oscillation, from their initial positions (blue) to their final positions (red). These drifters illustrate qualitatively different trajectories in the flow, including those that switch quadrants (dots), those that remain in a single eddy core (triangles), and those that spiral radially between the center and the boundary of a quadrant (squares).  In panel (B), the initial positions of 3000 randomly initialized drifters are colored according to their initial quadrant in the flow. The drifters maintain their color assignment in panel (C), showing how the unsteady eddy motion leads to mixing of the drifters after the 4 periods of horizontal oscillation. The east-west oscillation of the eddy field leads horizontal mixing of the flow. The resulting sCSC dendrogram is shown in panel (D), with every position occupied by all 3000 drifters plotted in black dots in the corresponding inset branch plot (note that drifter positions often appear as continuous black patches due to the high density of overlapping positions occupied by the drifters.) The width of each branch is proportional to the fraction of the states that it contains. The corresponding binary code of each branch is labeled in black text, and the number of trajectories associated with each node is labeled in red text.
The dendrogram is plotted to the seventh eigenvector, although labels below the fourth eigenvector are omitted for clarity.
The horizontal and vertical axes are measured in units of the parameter $z$, and the branches are plotted at 45-degree angles. We have visualized the first 7 eigenvectors for brevity of presentation.}
\label{fig:quadgyre}
\end{figure*}

\begin{figure*}[thb!]
\centering
\includegraphics[width=\textwidth]{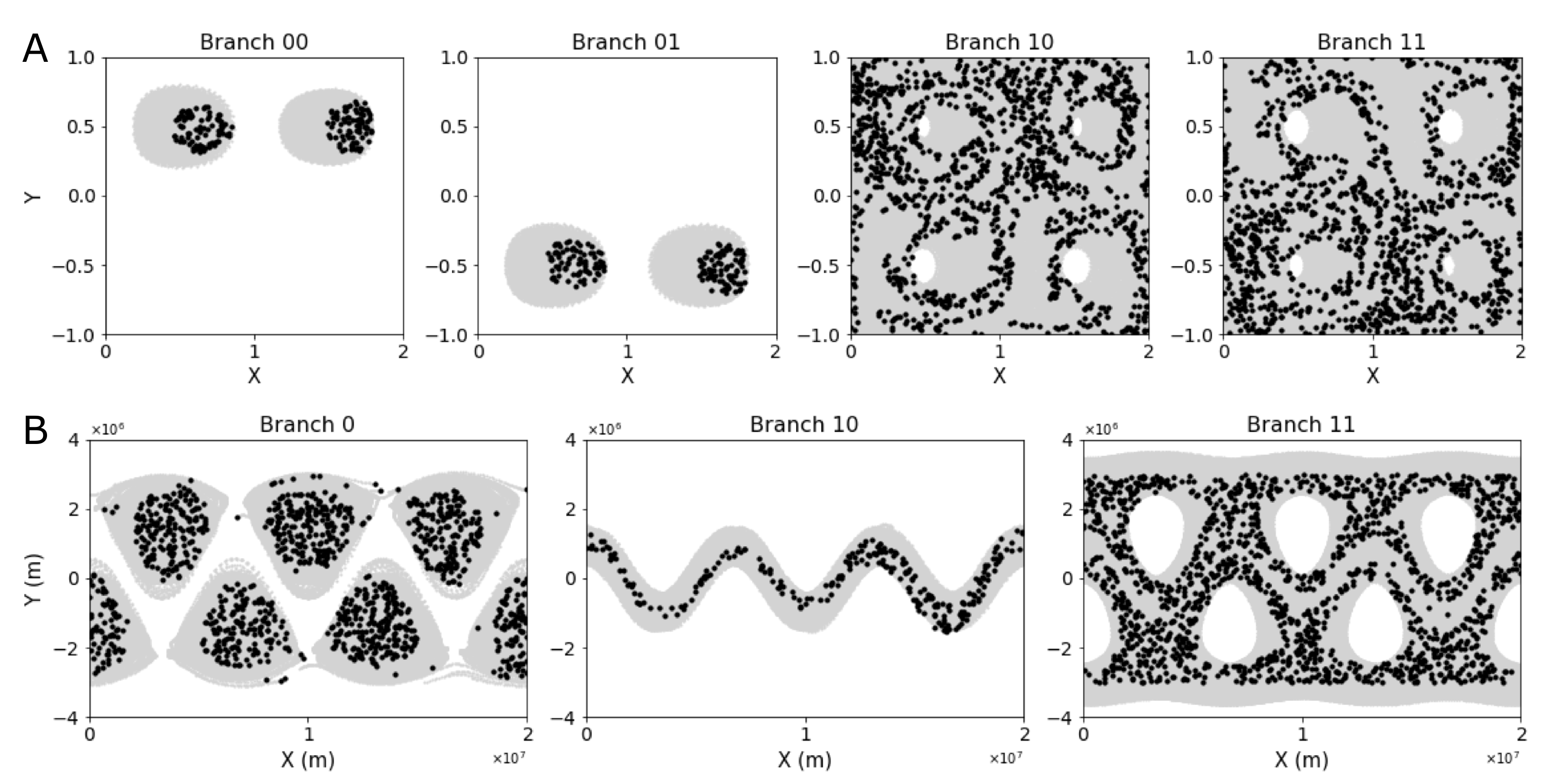}
\caption{Selected coherent structures from the quadruple-eddy and Bickley jet.
(A) Four selected branches from the quadruple-eddy analysis are visualized by plotting the particle trajectories belonging to a designated coherent structure (gray) and their starting positions (black).
Branches 00 and 01 show the top and bottom eddy cores, respectively, whereas branches 10 and 11 show the incoherent region. The latter two branches also display an imbalance between the top and bottom regions of the flow.
(B) Three selected branches from the Bickley jet analysis are visualized by plotting the particle trajectories belonging to a designated coherent structure (gray) and their starting positions (black).
Branch 0 shows the eddy cores, which do not mix. Branch 10 contains the meandering jet, and branch 11 accounts for the incoherent surroundings.
For both sets of plots, the starting positions show that the particles belonging to coherent structures at a given time point (in this case, the starting point) are more compactly located than the total space explored by their trajectories over time.
A particle found at a given instant in the space that is common among different coherent structures can therefore not be attributed to a coherent structure based on that time point alone.
Visually equivalent results can be produced from other time points.
}
\label{fig:structures}
\end{figure*}

\section*{Coherent structure identification from analytical geophysical flow simulations and empirical measurements}\label{sec:geophys}

\subsection*{Quadruple-eddy simulated ocean flow} \label{subsec:eddy}

A key challenge in geophysical fluid dynamics is to extract and characterize coherent fluid motions from sparsely sampled turbulent flows of air or water. 
The coherent structures, often manifested as eddies and jets, can dominate the transport of heat, salt, nutrients, and pollutants~\cite{Huhnetal_GRL12,HughesMiller_GRL17}.
Therefore, they can serve as the basis for low-order models that capture the salient physics~\cite{Kaiseretal_JFM14}, or as a template for data assimilation into large-scale weather forecasting models~\cite{Macleanetal_PhysicaD17}.
Turbulent flow structures in the ocean also impact the behavior and ecology of marine life~\cite{Stockeretal_Nature17}.

Distributed sensor networks such as the Argo collection of 3800 ocean drifters~\cite{Argo_00} sample the flow field in a Lagrangian sense, recording the properties of the water as each drifter is carried by the prevailing currents.
Here we demonstrate the capability of the sCSC method to extract coherent fluid structures from such collections of Lagrangian measurements.

To do so, we first apply sCSC to a common model of Lagrangian ocean drifters in a simplified flow field comprising only four eddies, the unsteady quadruple-eddy flow~\cite{AllshousePeacock_Chaos15,FroylandPadberg_Chaos15}.
While this model represents a simplification of the full physics, it is valuable due to its common use for the evaluation and comparison with existing methods to identify coherent structures~\cite{AllshousePeacock_Chaos15,FroylandPadberg_Chaos15,SchlueterKuck_JFM17,SchlueterKuck_Chaos17}.

As shown in Fig~\ref{fig:quadgyre}A, drifter trajectories within the two eddies at the upper-left and lower-right rotate clockwise, whereas trajectories within the other two eddies rotate counter-clockwise.
Simultaneous with this rotation, an east-west oscillation of the eddy field occurs, which causes exchange of drifters between the eastern and western eddies.
This exchange, which depends on the location and timing of the drifter release relative to the east-west oscillation cycle, is illustrated in the transition from initial drifter positions in Fig~\ref{fig:quadgyre}B to their final positions in Fig~\ref{fig:quadgyre}C. 

Each drifter trajectory represents a state of this fluid dynamic system, and the pairwise dissimilarity between each of the states is given by the standard deviation of the instantaneous distance between drifter positions at time $t_k$, $r_{ij}(t_k)$, divided by the average distance between each pair of drifters, ${\overline{r_{ij}}}$, for $T$ total time points~\cite{SchlueterKuck_JFM17}:

\begin{align}
a_{ij} = \frac{1}{\overline{r_{ij}}} \left[ \sum_{k=1}^T (\overline{r_{ij}}-r_{ij}(t_k))^2 \right]^{\frac{1}{2}}. \label{eqn:aij_fluids}
\end{align}

This measure anticipates that coherent structures will comprise drifters whose relative positions do not vary as the flow evolves, leading to a small values of the pairwise dissimilarity measure (i.e.~a small standard deviation) within each cluster.
By contrast, pairs of drifters that straddle the boundary between coherent structures can experience exponential separation over time and a correspondingly large standard deviation of their instantaneous separation~\cite{HallerYuan_PhysicaD00}.

Without requiring the specification of the number of eddies, the sCSC method reveals a clear, physically interpretable structure for this complex flow (Fig~\ref{fig:quadgyre}D).
The primary bifurcation of the flow is between trajectories that remain in the eddy cores of their original quadrant (branch 0) and trajectories that do not (branch 1).
The trajectories of branch 0 are then further subdivided into trajectories that remain within eddy cores in the northern half of the flow (branch 00) and those that remain within eddy cores in the southern half of the flow (branch 01), reflecting the absence of north-south drifter exchange.
Finally, the trajectories associated with the individual quadrants are identified at the level of the third bifurcation (e.g.~branch 000 shown in Fig~\ref{fig:quadgyre}D inset, as well as branches 001, 010, and 011 for the other three individual quadrants, not shown in inset). 
An additional visualization of the major coherent structures identified---namely, branches 00, 01, 10, and 11 in Fig~\ref{fig:quadgyre}D---is presented in Fig~\ref{fig:structures}A.

Whereas the application of $k$-means clustering or other conventional tools would require \textit{a priori} guidance to determine that four independent structures exist in branch 0 (i.e.~one eddy per quadrant)~\cite{Hadjighasemetal_Chaos17}, this result is revealed naturally by the sCSC dendrogram, as further bifurcations after branch 000 do not produce additional coherent states; all of the trajectories that remain together after the third bifurcation remain together after subsequent bifurcation.

To be sure, the presence of the four eddy cores can also be revealed by a contour map of the largest finite-time Lyapunov exponent (FTLE) corresponding to the quadruple-eddy velocity field (see Figs.~2 and~4 in Ref.~\onlinecite{SchlueterKuck_JFM17}).
The key advantage of the sCSC approach is that a similar result can be achieved with two orders-of-magnitude less data:~Schlueter-Kuck and Dabiri showed in Ref.~\onlinecite{SchlueterKuck_JFM17} that the FTLE gradient calculation is well-posed when the number of drifters is on the order of $10^5$
By contrast, the same cores can be identified by as few as 300 drifters using the present method, and the cores can be identified as long as drifters are present in the cores over timescales longer than the eddy turnover time.  

The structure of branch 1 is less well organized and reflects the chaotic advection of trajectories that spiral radially within a quadrant and/or switch quadrants in the unsteady flow. 
Nonetheless, the dendrogram structure does indicate geometric symmetries within the chaotic motions, such as a preference for three quadrants among the trajectories in branches 110 and 111; and a more constrained preference for two quadrants exists at branch 1110.
A general observation is that geometric symmetries appear as balanced dendrogram bifurcations.
This is in contrast to the structure of random noise, which is characterized by a trivial sCSC dendrogram with a single branch that contains nearly all of the states and a splintering of a small number of fully-converged states at each level of the dendrogram (see Fig.~\ref{S1_Fig}).

\begin{figure*}[thb!] 
\centering
\includegraphics[width=\textwidth]{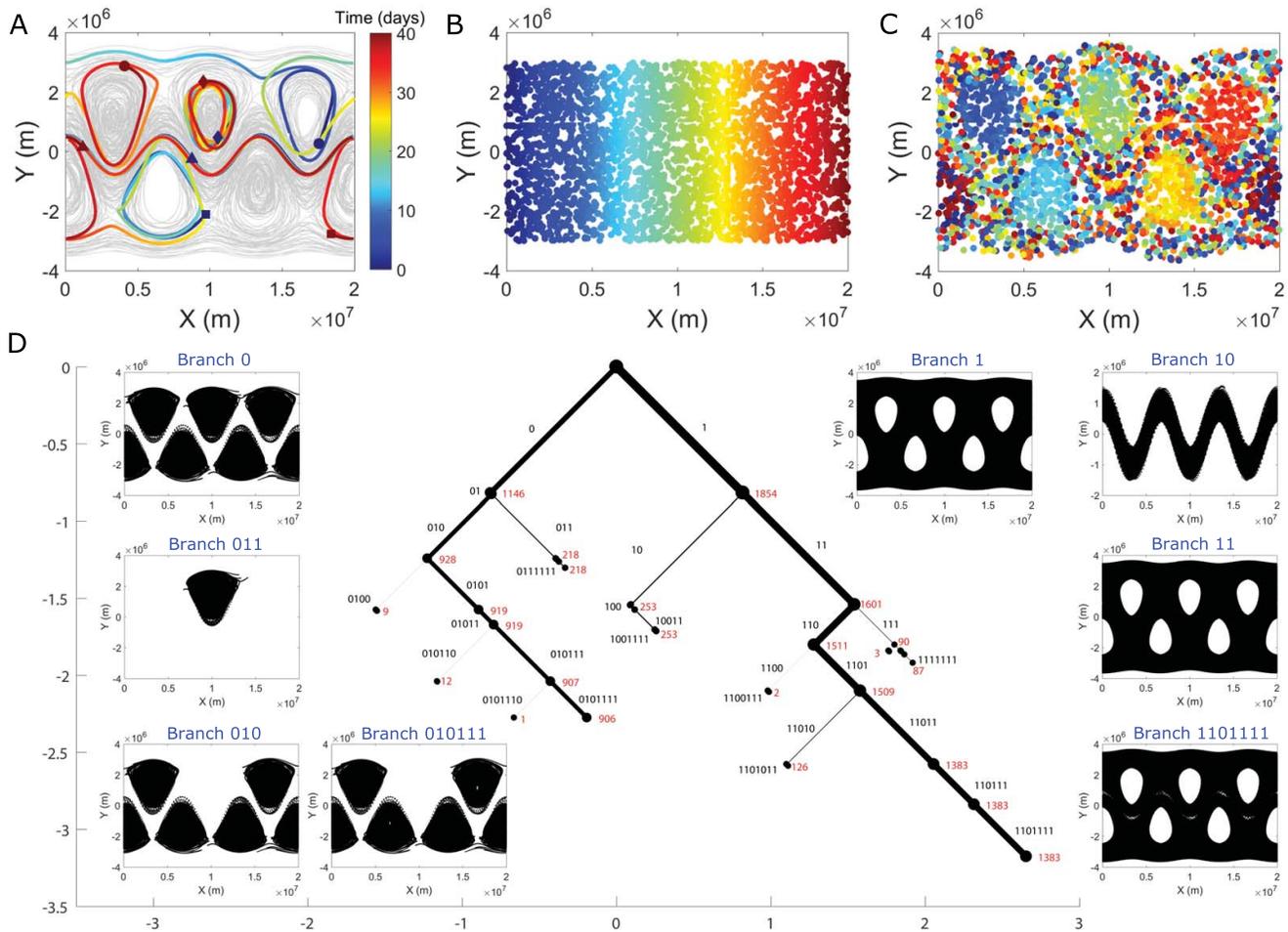}
\caption{Bickley jet atmospheric flow model.
(A) Trajectories of 75 selected Lagrangian particles randomly initialized in the flow (gray). The trajectories of 4 particles are highlighted for a 40-day integration period, from their initial positions (blue) to their final positions (red). These particles illustrate qualitatively different trajectories in the flow, including those the remain in a single flanking eddy (diamonds), those that pass between multiple eddies (dots and squares), and those in the meandering jet (triangles). In panel (B), the initial positions of 3000 particles are colored according to their position along the east-west axis of the flow. The particles maintain their color assignment in panel (C), showing how the unsteady jet and eddy motions lead to mixing of the particles after 40 days. A periodic boundary condition is applied in the east-west direction. The resulting sCSC dendrogram is shown in panel (D), with every position occupied by all 3000 particles plotted in black dots in the corresponding inset branch plot (note that particle positions often appear as continuous black patches due to the high density of overlapping positions occupied by the particles). The width of each branch is proportional to the fraction of the states that it contains. The corresponding binary code of each branch is labeled in black text, and the number of trajectories associated with each node is labeled in red text. The dendrogram is plotted to the seventh eigenvector, although many of the labels are omitted for clarity. The horizontal and vertical axes are measured in units of the parameter $z$, and the branches are plotted at 45-degree angles.
We have visualized the first 7 eigenvectors for brevity of presentation.}
\label{fig:bickley}
\end{figure*}

\subsection*{Bickley jet simulated atmospheric flow} \label{subsec:bickley}

A more complex geophysical flow model is the Bickley jet, which serves as a common model for zonal jets in the atmosphere\cite{Rypinaetal_JAS07}.
This flow is composed of a central meandering jet as well as flanking eddies that are periodic along the east-west axis (Fig~\ref{fig:bickley}A).
The sCSC dendrogram corresponding to this flow (for the same dissimilarity measure as the quadruple-eddy flow, equation~\eqref{eqn:aij_fluids}) is similarly effective in extracting the salient coherent features (Fig~\ref{fig:bickley}D).
The flanking eddies are identified in branch 0.

However, a key difference from the previous quadruple-eddy example is that the individual eddies are largely indistinguishable from one another.
This result reflects the homogeneity of fluid dynamics within the flanking eddies, which was not present among the trajectories in the quadruple-eddy flow (contrast e.g. Figs.~\ref{fig:quadgyre}C and~\ref{fig:bickley}C).
A notable exception is the eddy located at the meridional axis of symmetry, i.e.~branch 011.
An additional analysis tracking the distance of particles from the eddy cores showed that fewer particles belonging to this center eddy travel past a given contour threshold during the simulated time-series than particles from other eddies.
Branch 1 of the Bickley jet dendrogram collects those trajectories that are not associated with the flanking eddies. A subset of those trajectories, namely branch 10, is the meandering zonal jet. The remaining trajectories (branch 11) form a chaotic background flow that is robust to further bifurcation.
These three coherent structures are further visualized in Fig~\ref{fig:structures}B.

The sCSC structure of both of these simulated geophysical flows can be exploited to create low-order models of the governing fluid transport processes, without the need for \textit{ad hoc} assumptions regarding the number of coherent structures present.
Because similar results can be achieved despite significant missing or noisy data (see Ref.~\onlinecite{SchlueterKuck_JFM17}), the inherently limited data collection that can be achieved in the ocean and atmosphere can be more effectively leveraged to potentially improve the accuracy of weather forecasting, for example\cite{Macleanetal_PhysicaD17}. Hence, the sCSC method can be a powerful tool for both very large and very sparse datasets.

\subsection*{Empirical measurement of vortex ring formation and entrainment} \label{sec:vortex}

Vortex ring formation is a prominent phenomenon in engineered and biological systems as diverse as aerodynamic flow control, animal swimming, and the human cardiovascular system~\cite{KruegerGharib_POF03,Dabiri_ARFM09,Gharibetal_PNAS06}.
The growth and dynamics of vortex rings are dictated by the extent to which they entrain surrounding fluid~\cite{Dabiri_JFM04}.
Moreover, knowledge of the precise region of the flow that is ultimately entrained by a forming vortex ring can be used to predict how a vortex delivers mass, momentum, and energy to the surrounding flow.
For example, pathological vortex ring formation in the human left ventricle has been shown to provide an effective diagnostic of heart failure~\cite{Gharibetal_PNAS06}.
Despite the importance of vortex ring entrainment, methods to quantify the region of the flow impacted by vortex rings have shown limited success, particularly in cases for which the FTLE field cannot be calculated due to the sparsity of measurements.
Here, we demonstrate the ability of the sCSC technique to precisely identify the region of a flow that is entrained by a forming vortex ring---knowledge that has been previously inaccessible in cases where measurement data is sparse, such as when the flow is interrogated using non-invasive clinical methods such as ultrasound or magnetic resonance imaging. 

Vortex rings were formed in the laboratory using a piston-cylinder apparatus described in previous work~\cite{Schlueter_PRF16}. A motor-driven piston pushes water through a vertical hollow cylinder of diameter $D$ = 2.49 cm that is submerged in a tank with cross-sectional area of 61 cm by 61 cm and height of 91 cm. As the flow exits the cylinder at a nominal speed of 7 cm s$^{-1}$, the fluid boundary layer at the inner surface of the cylinder rolls up into a toroidal vortex ring, which propagates away from the cylinder via self-induction.

\begin{figure*}[htb!]
\centering
\includegraphics[width=\textwidth]{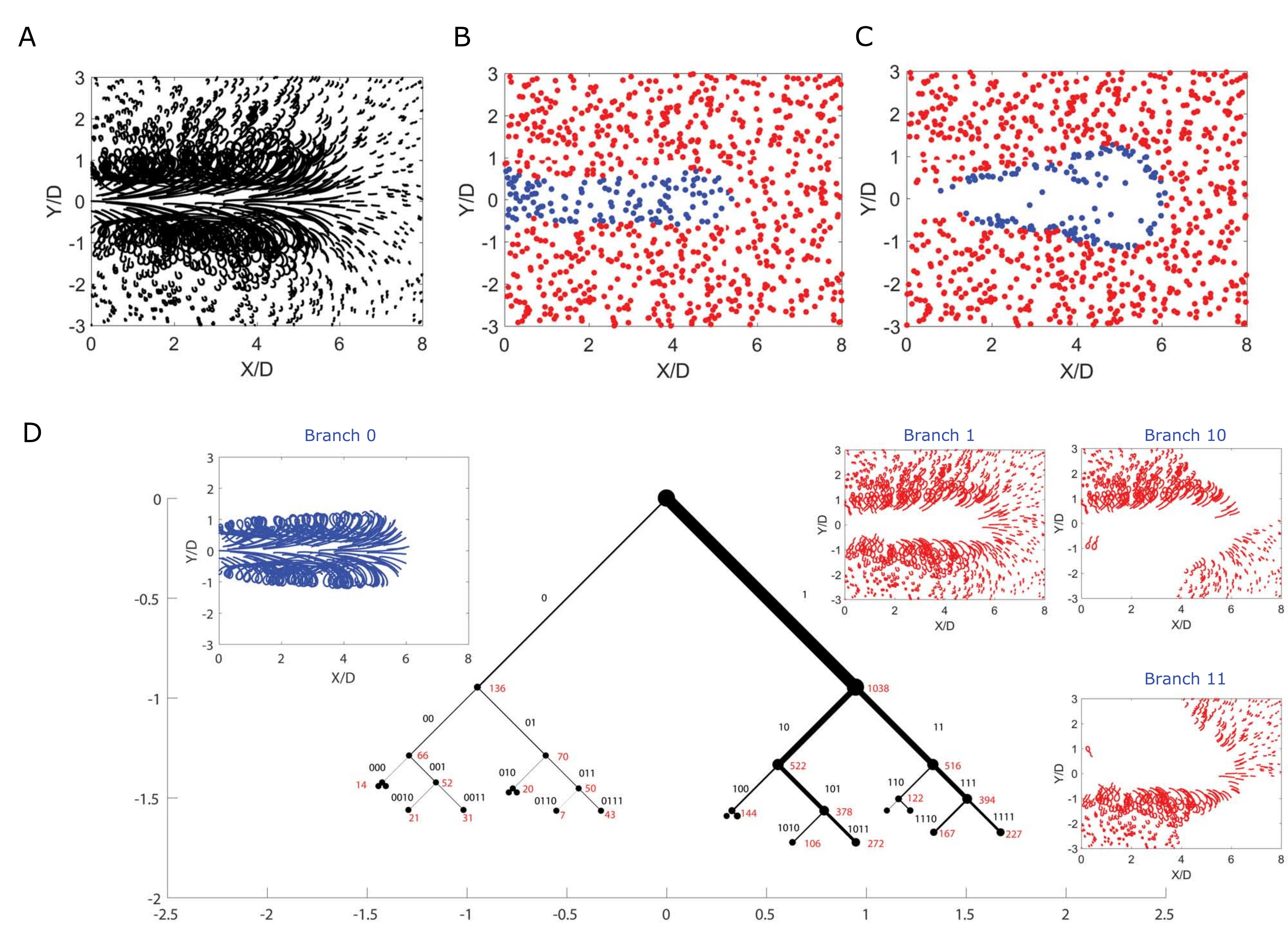}
\caption{sCSC analysis of vortex ring formation. (A) Trajectories of 1174 Lagrangian particles initialized in the flow. (B) Initial positions of the 1174 particles. Blue particles are those revealed by the sCSC analysis to be entrained by the vortex ring; red particles are not entrained. (C) Final positions of the same particles tracked in panel B. The sCSC dendrogram on which this analysis is based is shown in panel (D), with every position occupied by all 1174 particles plotted in the corresponding inset branch plot. The width of each branch is proportional to the fraction of the states that it contains. The corresponding binary code of each branch is labeled in black text, and the number of trajectories associated with each node is labeled in red text. The dendrogram is plotted to the fourth eigenvector, although many of the labels are omitted for clarity. The horizontal and vertical axes are measured in units of the parameter $z$, and the branches are plotted at 45-degree angles. We have visualized the first 4 eigenvectors for brevity of presentation, but this does not affect the model results.}
\label{vorscsc}
\end{figure*}

A set of 1174 fluid particle trajectories in the domain encountered by the vortex ring were analyzed using the present sCSC method and the dissimilarity measure in equation~\eqref{eqn:aij_fluids} to identify regions of the ambient flow that were entrained by the vortex ring. As illustrated in Fig.~\ref{vorscsc}A, it is impossible to determine which fluid particles have been entrained by the vortex ring based on visual inspection of the trajectories alone.
A comparison FTLE analysis performed by Schlueter-Kuck and Dabiri on 30,500 advected particles (see Ref.~\onlinecite{SchlueterKuck_JFM17}, Fig.~9) showed that 1174 trajectories are not sufficiently close to one another to compute the FTLE field, because the required gradient calculations are not well-posed for sparse trajectories.
The alternative use of existing techniques based on heuristics, such as $k$-means or the spectral eigengap, rely on knowledge of the number of eddies to guide clustering; in the present case, it is not known \textit{a priori} how many structures comprise the flow.

The sCSC dendrogram (Fig.~\ref{vorscsc}D) avoids the need for explicit determination of the number of eddies, as it unambiguously identifies the fluid particles entrained by the vortex ring as those belonging to branch 0. Branch 1 identifies all other particles and further bifurcations of that branch reveal underlying geometric symmetries, as in Branch 1 of the quadruple-eddy flow in Fig.~\ref{fig:quadgyre}D. 

Plots of the initial and final positions of the fluid particles in Fig.~\ref{vorscsc}B and C show that the fluid entrained by the vortex ring occupies a well-defined region in the immediate path of the vortex ring, a result that is consistent with intuition but that can now be characterized quantitatively for the first time.
The void created by the evolution of the blue particles from Fig.~\ref{vorscsc}B to Fig.~\ref{vorscsc}C is filled by the fluid ejected from the cylinder.
The entrained blue particles ultimately occupy positions around the vortex ring that are consistent with the FTLE analysis in Ref.~\onlinecite{SchlueterKuck_JFM17}.
This provides another demonstration of the interpretability of the sCSC results:~notably, these result have been achieved without any of the ad hoc assumptions required by existing methods of entrainment quantification~\cite{Dabiri_JFM04,Olcay_EIF08}.

\section*{Visualizing macrostate modeling of molecular dynamics} \label{sec:prg}

In this section we highlight the interpretability of the sCSC dendrogram for a high-dimensional dynamical dataset.
Specifically, we focus on an atomistic simulation of protein folding.
Whereas fluid dynamics datasets typical represent only a few spatiotemporal coordinates, atomistic molecular dynamics (MD) datasets can contain thousands of degrees of freedom with complex interrelationships.

While MD is resource-intensive, advances in simulation parameters~\cite{Wang_JPCB2017}, bespoke hardware~\cite{Shaw_Science10}, and distributed computing frameworks~\cite{Shirts_Science00}, have enabled MD analyses to yield insight into complex biophysical systems at biologically meaningful timescales~\cite{Husic_JACS18}. 
Thus, these simulations have the potential to uncover biophysical phenomena such as the misfolding mechanisms involved in a variety of diseases, stable configurations yet undiscovered by crystallography, and small molecule binding sites and kinetics for drug discovery.

However, without complementary analysis methods designed to communicate statistically rigorous and understandable conclusions resulting from such computational experiments, the benefits of advances in MD cannot be fully realized.
While many methods have been developed to perform these analyses~\cite{Husic_JACS18}, it remains a challenge to display their results in a meaningful way.
sCSC can be used to augment already-existing methods for analyzing MD simulations such that the results can be visualized and interpreted.

To demonstrate the use of sCSC to visualize an MD analysis, we use an ultralong MD simulation performed by Lindorff-Larsen \textit{et al}.~\cite{LindorffLarsen_Science11} of the folding and unfolding of Protein G, a $56$-residue protein expressed in streptococcal bacteria.
The simulation details are described in the Supporting Materials of Ref.~\onlinecite{LindorffLarsen_Science11}.
We use a Markov state model (MSM) analysis to define the states of the system, which is a discrete approximation to the Perron-Frobenius operator~\cite{schutte1999direct}.
This established mathematical framework codifies the system using a kinetic master equation~\cite{Husic_JACS18}.
The master equation takes the form of a stochastic transition probability matrix, in which each state of the system is identified by a probability distribution of transitioning to every other state.

After constructing a quantitatively accurate and optimized MSM~\cite{Noe_MMS13}, we are interested in clustering these state into a smaller number of interpretable ``macrostates'', since it is conceptually difficult to describe hundreds unique states in a physically interpretable way~\cite{Pande_Methods10}.

For our MSM, we found that 175 states optimally describes the system according to a variational evaluation 
(the MSM construction protocol is consistent with current best practices and is described in detail in the Methods).
Minimum variance clustering analysis (MVCA), an effective coarse-graining method for MSMs, has recently been developed by one of the authors and uses a pairwise information theoretic dissimilarity metric in order to group states into a smaller number of macrostates, namely, the Jensen-Shannon divergence between the probability distributions characterizing the rows of the MSM transition probability matrix~\cite{Husic_Draft17} (see also Methods equation~\eqref{eqn:js}).

\begin{figure*}[thb!]
\centering
\includegraphics[width=\textwidth]{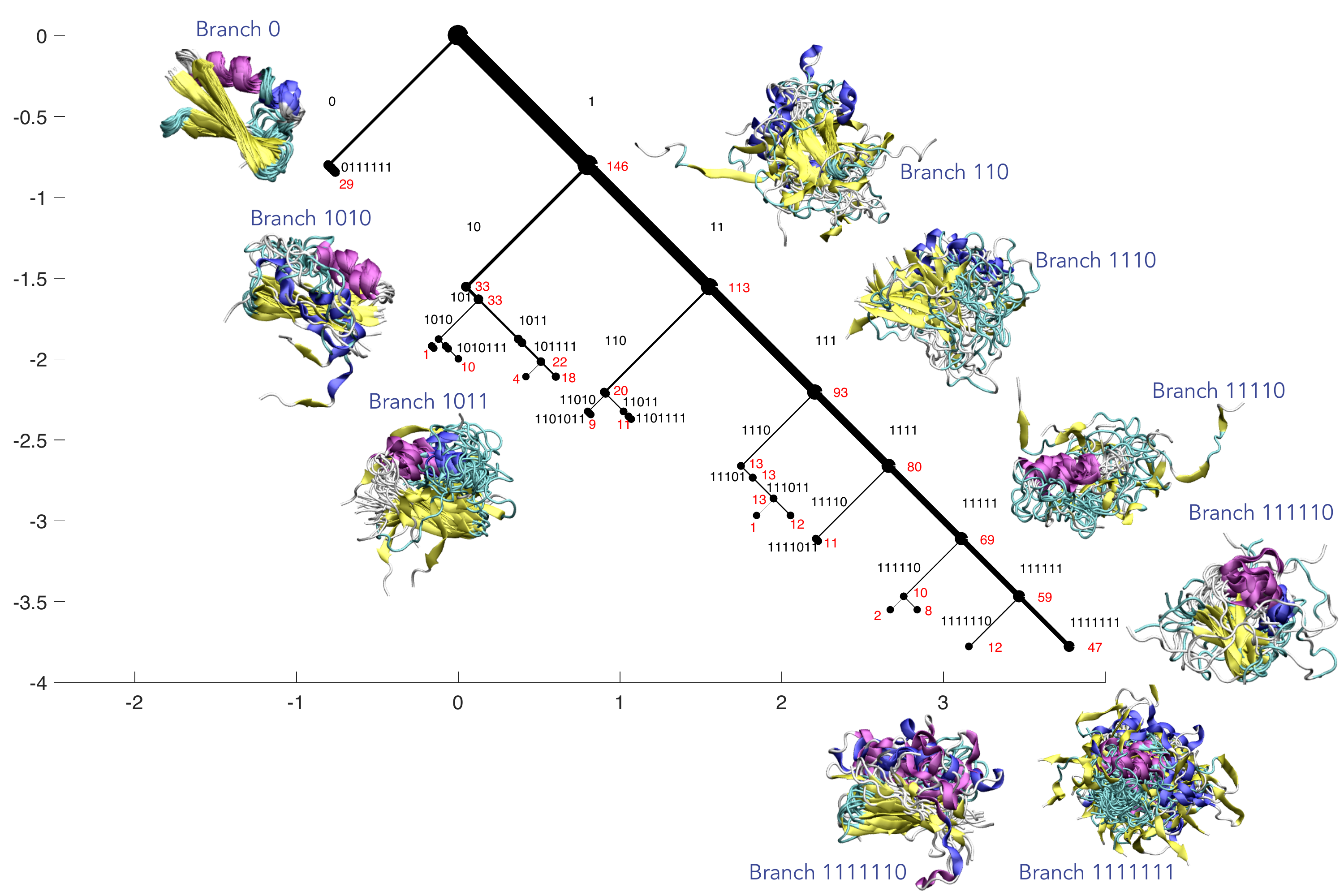}
\caption{Atomistic protein folding simulation.
The sCSC dendrogram for protein folding simulations of Protein G isolates the folded state and eight other macrostates that characterize the folding process.
Each collective structure represents one sample drawn from each MSM state contained in the macrostate.
The 9 macrostates contain all 175 original states, and are distinguished by consistency in secondary structure:~pink regions represent $\alpha$-helices, blue represents $3_{10}$ helices, and yellow represents $\beta$-sheet regions. Branch 0, which contains the folded macrostate, is the most ordered and is separated first. 
Branch 1111111 represents the most disordered macrostate. The other branches represent macrostates with varying degrees of order and secondary structure elements. 
The width of each branch is proportional to the fraction of the states that it contains.
The corresponding binary code of each branch is labeled in black text, and the number of MSM states associated with each node is labeled in red text.
The horizontal and vertical axes are measured in units of the parameter $z$.
We have visualized the first 7 eigenvectors for brevity of presentation.}
\label{fig:prg}
\end{figure*}

By using the same pairwise dissimilarity metric as MVCA, the ${175 \choose 2}$ state adjacencies can be input into the sCSC algorithm to produce a visualization of a set of macrostates in the protein folding dataset, which is displayed in Fig.~\ref{fig:prg}.
Nine branches of the sCSC dendrogram are depicted by sampling one structure from each original MSM state contained in that branch.
Since the nine depicted branches contain all 175 original MSM states, these branches can be interpreted as a possible set of system macrostates.
By superimposing a representative conformation from each MSM state and coloring the protein according to its secondary structure, we can visualize the MD trajectory by interpreting the sCSC groupings.

First, we note that the folded structure (branch 0) is identified in the first sCSC solution and is separated from the denatured, unfolded ensemble, which comprises the rest of the dendrogram (branch 1).
We see that the folded branch isolates a well-defined conformation with low variance across sampled conformations.
The incorporation of subsequent sCSC eigenvectors identifies groups of structures unified by their protein secondary structure features.
Various branches contain similar secondary structure elements (similar colors in the structure visualization in Fig.~\ref{fig:prg}), elucidating substructures exhibited during the folding of Protein G.
For example, branch 1110 contains $\beta$-sheet secondary structure (yellow), whereas branch 11110 contains noticeable $\alpha$-helical secondary structure (pink).
Branch 1111111 is the least coherent, containing the most unstructured states.
Summary statistics for each macrostate can be found in Table~\ref{S1_Table}.

In this example, we have chosen to highlight secondary structure changes so we can understand which secondary structure elements characterize different subprocesses within folding. We see that the yellow $\beta$-sheet secondary structure appears in several macrostates---often along with the blue $3_{10}$-helix, thought to be an intermediate structure during $\alpha$-helix formation~\cite{Sundaralingam_Science89}---which might indicate that the formation of the pink $\alpha$-helix is a rate-limiting step in the folding process. However, we could also choose to quantify and visualize macrostate contact maps, radii of gyration, or distance to various structures in order to gain complementary insight into the folding system.

For other dynamical processes characteristic of proteins, such as conformational change, allostery, and drug binding, we might choose to visualize parameters related to specific sites of interest or observables that can be probed experimentally.
The choice of how to describe the macrostates is independent of the clustering process, but the depictions or statistics that enable the best interpretation of the system will depend on the dynamics of interest. 

The sCSC dendrogram analysis offers advantages in the interpretation of high-dimensional MD datasets after an adequate kinetic model has been constructed.
Utilizing the pairwise state adjacency from this kinetic model for an sCSC analysis produces a hierarchical representation of structural motifs according to the extent of their dissimilarity, which provides insight into
the protein conformations that characterize subprocesses within folding.
As in the fluid dynamics examples in the previous section, truncating the dendrogram when bifurcations are unoccupied produces an objective way to visualize the converged clusters.
Finally, the generation of orthogonal sCSC solutions enables orthogonal dynamical processes in the protein folding simulation to be incorporated in analogy to the simple model in Fig~\ref{fig:scheme}.
We anticipate that this type of interpretable visualization will be useful for communicating the results obtained from high-dimensional datasets.

\section*{Discussion}

The present approach addresses the previously stated challenges with common clustering algorithms:~it does not require a choice of cluster number or dendrogram cutting, it leverages the concept of dissimilarity in a computationally tractable way, and it maintains an interpretable hierarchical relationship among splittings.

Perhaps the most important advantage of this approach relative to commonly used tools is that the number, shape, and size of clusters in the data emerges naturally from the sCSC dendrogram rather than being specified \textit{a priori}.
As the set of eigenvectors that is included in the analysis is increased to include those associated with lesser eigenvalues, the number of unoccupied binary codes generally increases. 
This is because progressively fewer groups of states that survived the preceding orthogonal partitions will be subsequently separated at lower levels of the dendrogram.
In this way, the set of clusters in the dataset is revealed to be those branches of the dendrogram structure whose shape converges as the number of eigenvectors included in the analysis increases.
The sCSC dendrogram indicates not only the number of these converged clusters but also the relative strength of the partitions between clusters, via the length of the connecting branches in $z$-space.

While sCSC conceptually resembles divisive hierarchical clustering, the number of possible divisions in the latter scales as $2^{c-1}-1$ with the number of clusters $c$, which is generally not feasible for large $c$ unless the initial dataset is sparse~\cite{Boley_DMKD98}.
sCSC scales in the same way as agglomerative hierarchical clustering, requiring a computationally nontrivial but tractable calculation of ${n \choose 2}$ dissimilarity values for $n$ initial data points.
However, unlike agglomerative clustering---which also requires the calculation of ${n \choose 2}$ dissimilarities---small differences between states at lower levels of the sCSC dendrogram have no impact on the clusters that form at higher levels, as the top-down approach begins by using the most significant partitions indicated by the eigenvectors associated with the largest eigenvalues.

When applying sCSC, domain knowledge should inform selection of an appropriate dissimilarity measure, but \textit{ad hoc} and \textit{a priori} assumptions about the structure of the data itself are not needed.
While we have demonstrated sCSC only for simulated physical systems,
we anticipate that these features will make sCSC a powerful tool for interrogating both new and longstanding research problems, including those in fields where the underlying processes are less accessible, such as genomics and neuroscience.
For example, genetic ancestries can potentially be clustered on the basis of the sCSC structure that emerges from the dissimilarity of single-nucleotide polymorphisms (SNPs) among individuals within a population.
In the latter case, differences in neuronal activation can be amplified using sCSC to identify emergent functions that involve coordination of spatially distant neurons.
These and other applications can be pursued immediately given the tools developed here.

\section*{Methods}
\subsection*{Quadruple-eddy ocean flow model}
 The velocity field of the quadruple-eddy ocean flow model is given by,

\begin{eqnarray*}
\frac{dx}{dt} & =&-\pi A\sin(\pi f)\cos(\pi y)\\
\frac{dy}{dt} & =&-\pi A\cos(\pi f)\cos(\pi y)(2ax+b),
\end{eqnarray*}

\noindent{}where $x = [0, 2]$ and $y = [-1, 1]$ are the dimensionless east-west and north-south spatial coordinates (i.e.~normalized by the quadrant side length), $t$ is time in dimensionless units, and
\begin{eqnarray*}
a =\epsilon\sin(\omega t)\\ 
b =1-2\epsilon\sin(\omega t)\\ 
f =ax^2+bx.
\end{eqnarray*}

\noindent{}In the present unsteady implementation of the model, $A=0.1$, $\epsilon=0.1$, and $\omega=2\pi/10$.  3000 Lagrangian drifters were randomly initialized in the domain and advected in the flow using a fifth-order Runge-Kutta integration scheme. The duration of advection was 40 dimensionless time units, corresponding to 4 periods of horizontal oscillation of the flow.

\subsection*{Bickley jet atmospheric model}
The Bickley jet flow is given by the streamfunction $\psi=\psi_0+\psi_1$, where,
\begin{eqnarray*}
\label{eq:bickley_eqn1a}
\psi_0 & =&c_3y-UL\tanh\left(y/L\right)\\
\label{eq:bickley_eqn1b}
\psi_1 & =&UL\textrm{ sech}^2\left(y/L\right)\sum_{n=1}^3\epsilon_n\cos\left(k_n\left(x-\sigma_nt\right)\right).
\end{eqnarray*}

In the present study, we use similar parameter values as in Ref.~\onlinecite{Hadjighasemetal_PRE16}: $U=62.66$ ms$^{-1}$, $L=1770$ km, $k_n=2n/r_0$, $c=[0.1446U$, $0.205U$, $0.461U]$,  $\sigma=c-c(3)$, and $\epsilon=[0.0075$, $0.15$, $0.3]$, and the flow is computed on the interval $x=[0$, $20\times10^6]$ m, $y=[-3\times10^6$, $3\times10^6]$ m, over the time interval $t=[0$, $40]$ days, divided into 601 discrete time steps.  The flow was treated as periodic in $x$.  3000 Lagrangian fluid particles were randomly initialized in the domain and advected in the flow using a fifth-order Runge-Kutta integration scheme.

\subsection*{Markov state models}
Markov state models (MSMs) are a kinetic master equation framework for describing and analyzing time-series data such as molecular dynamics (MD) simulations by approximating the continuous Perron-Frobenius operator using a discrete transition probability matrix~\cite{schutte1999direct}.
A MSM requires partitioning the phase space explored by a system into discrete states (henceforth ``microstates''), and is represented by a transition probability matrix defined for a Markovian lag time $\tau$ at which transitions between the microstates are independent of the history of the system. For protein folding analyses, phase space (positions and velocities) is conventionally approximated by conformation space (positions), and states are chosen according to an objective optimization protocol, in this case a variational principle~\cite{Noe_MMS13}. The Markovian lag time chosen to analyze a system must be long enough for memoryless inter-state transitions, but short enough to resolve dynamics; for protein folding dynamics, lag times on the order of tens of nanoseconds are typical.

The MSM transition probability matrix is constrained to be stochastic, symmetric with respect to a stationary distribution, ergodic, and aperiodic. It can thus be decomposed into eigenvalues and eigenvectors, $T(\tau)\lambda = \psi\lambda$, where the eigenvalues are on the unit interval $|\lambda_i| \leq 1$ and the highest eigenvalue $\lambda_1 = 1$ is unique. The variational principle states that the sum of estimated eigenvalues is bounded from above by the sum of true eigenvalues; thus many state decompositions can be tested according to the sum of a set number of eigenvalues for a set Markovian lag time and the state decomposition resulting in the highest sum of approximated eigenvalues can be chosen for further analysis.

The MSM for the simulation analyzed in this work was constructed according to the protocol used in Ref.~\onlinecite{Husic_JCTC17} for a set lag time of 50~ns according to a previous analysis for the same system performed in Ref.~\onlinecite{Beauchamp_PNAS12}. First, the Cartesian coordinates from the raw simulation data are transformed into the sines and cosines of the $\phi$ and $\psi$ side chain dihedral angles for each amino acid of the protein. Next, the vector of dihedrals is again transformed using time structure-based independent component analysis (tICA)~\cite{Schwantes_JCTC13} with a tICA lag time of 50~ns wherein each tICA solution vector was weighted according to its associated eigenvalue~\cite{Noe_JCTC16}. Then, mini-batch $k$-means was used to cluster the simulation frames according to their weighted tICA representations for 265 different numbers of cluster centers randomly chosen between 10 and 5000. Finally, a MSM was constructed on each $k$-means state decomposition in which the transition probability matrix is obtained using a maximum likelihood estimator of the data such that detailed balance is achieved.
For each model, five MSMs were fit to a randomly chosen half of the data and then applied to the other half of the data, and the latter was used to sum the first 50 MSM eigenvalues as that model's score. The winning model was chosen to be the one that achieved the single maximum score from parameter sets with mean scores within one standard deviation of the maximum mean score. For our analysis of 265 different microstate numbers, the best model according to this variational analysis had 175 microstates and was used for analysis in the main text.

\subsection*{Coarse-graining MSMs with MVCA}
Minimum variance clustering analysis (MVCA) was recently published by one of the authors as an algorithm for coarse-graining an MSM transition probability matrix into a smaller number of macrostates by grouping the original microstates~\cite{Husic_Draft17}.
MVCA achieves a coarse-grained model by using agglomerative hierarchical clustering with Ward's minimum variance method~\cite{ward1963hierarchical} to cluster the microstates, where the pairwise dissimilarity between microstates is quantified using an information theoretic measure between the probability distribution characterized by the corresponding row of the MSM transition matrix.

If two microstates are defined by transition probability distributions $P$ and $Q$, their pairwise dissimilarity can be written using the Jensen-Shannon divergence~\cite{Lin_IEEE91},

\begin{equation} \label{eqn:js}
\textup{div}_{JS}(P||Q) = \frac{1}{2}\sum_{i}P_i\log{\frac{P_i}{M_i}} + \frac{1}{2}\sum_{i}Q_i\log{\frac{Q_i}{M_i}}
\end{equation}

\noindent{}where $M$ is the elementwise mean of $P$ and $Q$, and each term is the Kullback-Leibler divergence to the mean.
We quantify the dissimilarity between microstates using the square root of equation~\ref{eqn:js}~\cite{Endres_IEEE03,Husic_Draft17}.

From this set of pairwise similarities, MVCA goes on to cluster the microstates using agglomerative hierarchical clustering with Ward's method. In the analysis presented in this work, the set of pairwise dissimilarities is instead used to construct the adjacency matrix for sCSC.

\section*{Data availability}

The three fluid mechanics datasets and all adjacency matrices used to create the models in this work are available on github at \url{https://github.com/brookehus/sCSC}.
This repository also contains example MATLAB and Python codes, including Jupyter notebook tutorials.
The all-atom molecular dynamics simulations of Protein G were previously published in Ref.~\onlinecite{LindorffLarsen_Science11}, and the trajectories are available at no cost for non-commercial use through contacting \texttt{trajectories@deshawresearch.com}.

\section*{Acknowledgments}
The authors are grateful to Muneeb Sultan, Jared Dunnmon, Nicole Xu, and the referees for insightful manuscript feedback and to D.~E.~Shaw Research for providing the Protein G dataset.
This work was supported by the U.S.~National Science Foundation and by the Department of Defense (DoD) through the National Defense Science \& Engineering Graduate Fellowship (NDSEG) Program.

\clearpage

\onecolumngrid

\renewcommand{\thefigure}{S\arabic{figure}}
\renewcommand{\thetable}{S\arabic{figure}}
\setcounter{figure}{0}

\begin{figure*}
\centering
\includegraphics[width=\textwidth]{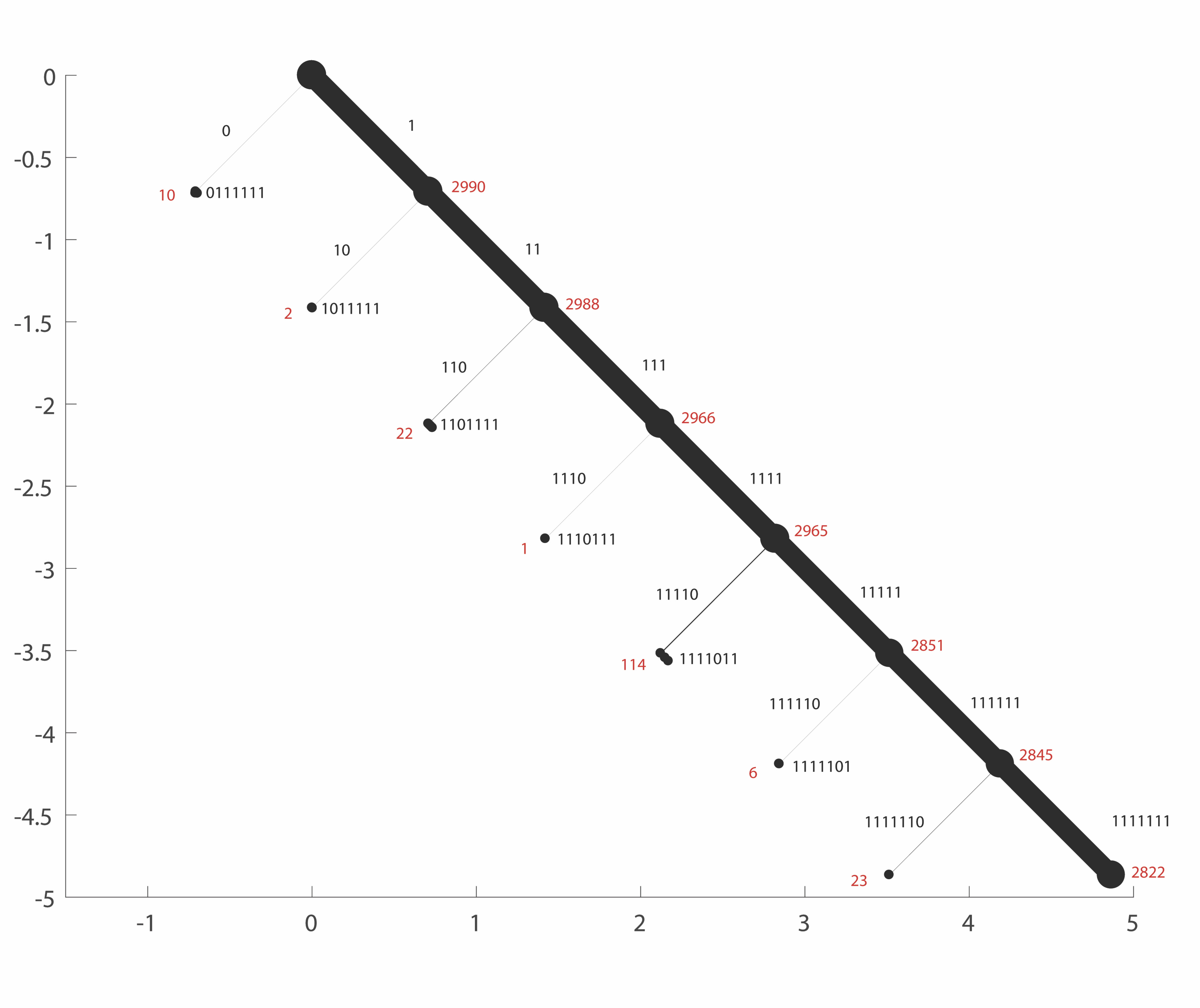}
\caption{sCSC dendrogram applied to random data.
To evaluate the sCSC dendrogram structure resulting from random noise, an adjacency matrix was constructed based on 3000 two-dimensional trajectories whose instantaneous positions over 2000 time steps were selected randomly from uniform distributions over the spatial coordinate intervals $x = (0, 1)$, $y = (0, 1)$.
These states were analyzed using pairwise dissimilarity based on the normalized standard deviation.
The result is a single main branch with a small splintering of trajectories at each eigenvector level. The splintering at each level converges throughout the seven eigenvectors included in the analysis.
The width of each branch is proportional to the fraction of the states that it contains. The corresponding binary code of each branch is labeled in black text, and the number of trajectories associated with each node is labeled in red text.}
\label{S1_Fig}
\end{figure*}

\begin{table*}
  \begin{tabular}{lccccc}
    \hline
    Branch & Population & Pairwise RMSD (\AA) & \% $\alpha$-helical & \% $\beta$-sheet & Aligned residues\\
    0 		& 29 & $2.35  \pm 0.34 $
    			 & $25.0 \pm 3.2 $ 
    			 & $43.7 \pm 2.7 $ 
                 & 1--56  \\
    1010 	& 11 & $7.78  \pm 2.04 $
    			 & $20.3 \pm 6.1 $ 
                 & $34.1 \pm 4.0 $ 
                 & 2--41  \\
    1011 	& 22 & $9.63  \pm 2.27 $
    			 & $ 17.5 \pm 7.7 $ 
                 & $ 31.1 \pm 5.6 $ 
                 & 10--40  \\
    110 	& 20 & $11.95 \pm 1.96$
    			 & $ 4.0 \pm 4.1 $ 
                 & $ 36.1 \pm 6.9 $ 
                 & 8--24  \\
    1110 	& 13 & $12.01 \pm 2.14$
    			 & $ 5.2 \pm 8.4 $ 
                 & $ 32.1 \pm 5.9 $ 
                 & 1--39  \\
    11110 	& 11 & $12.06 \pm 1.93$
    			 & $ 24.0 \pm 10.4 $ 
                 & $ 5.7 \pm 8.9 $ 
                 & 29--45  \\
    111110 	& 10 & $11.29 \pm 1.85$
     			 & $ 8.7 \pm 7.0 $ 
                 & $ 32.0 \pm 10.8 $ 
                 & 11--37  \\
    1111110 & 12 & $10.59 \pm 2.11$
    			 & $ 12.9 \pm 6.3 $ 
                 & $ 33.0 \pm 10.8 $ 
                 & 1--35  \\
    1111111 & 47 & $11.28 \pm 2.40$
    			 & $ 14.1 \pm 8.6 $ 
                 & $ 28.7 \pm 10.2 $ 
                 & 1--37 \\    
    Total	& 175 & $10.76 \pm 2.85$
    			 & $ 15.1 \pm 9.9 $ 
                 & $ 31.9 \pm 11.3 $ 
                 & -  \\   
    \hline
  \end{tabular}
      \caption{Macrostate statistics for Protein G.
Summary statistics for the nine macrostates identified from the sCSC model of Protein G. The first column identifies the branch label in Fig.~\ref{fig:prg}. The second column shows the number of the original 175 MSM states contained in the macrostate. For the next three columns, statistics are gathered using the one sampled state from each macrostate that is used for visualization in Fig.~\ref{fig:prg}. The third column reports the average $\pm$ one standard deviation pairwise RMSD over all atoms for all pairs of original MSM states within the same macrostate (e.g.~the average RMSD for Branch 0, which contains 29 microstates, each of which is represented by one sampled structure from each original MSM state, is the average of the pairwise RMSD for $(\genfrac{}{}{0pt}{}{29}{2}) = 406$ possible pairs). The fourth and fifth columns report the average percentage $\pm$ one standard deviation of the $\alpha$-helical and $\beta$-sheet secondary structure for each original MSM structure sample in the macrostate according to the simplified dictionary of protein secondary structure (DSSP) protocol\cite{Kabsch_Biopolymers83} implemented in the MDTraj\cite{McGibbon_BiophysJ15} software package. The last column indicates which Protein G residues were aligned to create the superpositions illustrated in Fig~\ref{fig:prg} for each branch.}
\label{S1_Table}
\end{table*}

\end{document}